\documentclass[runningheads]{llncs}

\usepackage{eccv}

\usepackage{eccvabbrv}

\usepackage{graphicx}
\usepackage{booktabs}

\usepackage[accsupp]{axessibility}  

\usepackage{hyperref}

\usepackage{orcidlink}

\begin{document}

\title{\textcolor{teal}{DEAR}: \textcolor{teal}{D}epth-\textcolor{teal}{E}nhanced \textcolor{teal}{A}ction \textcolor{teal}{R}ecognition}

\titlerunning{DEAR}

\author{Sadegh Rahmaniboldaji\inst{1}\orcidlink{0000-0002-7220-0234} \and
Filip Rybansky\inst{2} \and
Quoc Vuong\inst{2} \and
Frank Guerin\inst{1}\orcidlink{0000-0003-1918-6311} \and
Andrew Gilbert\inst{1}\orcidlink{0000-0003-3898-0596}}

\authorrunning{S. Rahmaniboldaji et al.}

\institute{University of Surrey, Guildford, UK \and
University of Newcastle, Newcastle, UK \\
\email{s.rahmani@surrey.ac.uk}\\
}

\maketitle

\begin{abstract}
    Detecting actions in videos, particularly within cluttered scenes, poses significant challenges due to the limitations of 2D frame analysis from a camera perspective. Unlike human vision, which benefits from 3D understanding, recognizing actions in such environments can be difficult. This research introduces a novel approach integrating 3D features and depth maps alongside RGB features to enhance action recognition accuracy. Our method involves processing estimated depth maps through a separate branch from the RGB feature encoder and fusing the features to understand the scene and actions comprehensively. Using the Side4Video framework and VideoMamba, which employ CLIP and VisionMamba for spatial feature extraction, our approach outperformed our implementation of the Side4Video network on the Something-Something V2 dataset. \textit{Our code is available at:} \textcolor{purple}{https://github.com/SadeghRahmaniB/DEAR}

\end{abstract}

\section{Introduction}
\label{sec:intro}

Substantial attention in computer vision has focused on Human Action Recognition (HAR) from videos because of its wide range of applications, including surveillance, human-computer interaction and content-based video retrieval. Despite significant advancements, recognising actions in cluttered scenes remains a challenging problem. This difficulty arises from the inherent limitations in analysing 2D frames captured from a single camera perspective, which often fails to represent real-world environments' three-dimensional (3D) nature.

Human vision can utilise 3D depth information to efficiently and robustly interpret people’s actions \cite{Vanrie2006, Wang2014}. Such information can be recovered from a combination of depth cues, such as optic flow and pictorial cues estimated from 2D frames \cite{brenner2018, Welchman2016}. Significantly, depth information can enhance perception of critical 3D spatial arrangements of people and objects \cite{Fairchild2024.03.16.585308, Michael2018}, segment actions and objects from background clutter \cite{Nakayama1989}, and complement other visual cues to differentiate actions under challenging viewing conditions \cite{Chainay2001, Humphrey1993, Jiang2008}.



Using 3D features, such as depth maps, in HAR systems may thus provide critical information about the spatial arrangement of objects within a scene, offering a more holistic understanding that strictly 2D approaches may miss \cite{liu2020attentiondistillationlearningvideo, kondratyuk2021movinetsmobilevideonetworks, li2022mvitv2improvedmultiscalevision}. Previous studies used different methods to incorporate depth maps as additional input to enhance HAR. These methods include combining depth maps with RGB data in CNN-based networks \cite{Garcia_2018_ECCV, 7732038, s20113305}, with inertial sensors such as accelerometers and gyroscopes  \cite{ahmad2020cnnbasedmultistagegated}, and with human posture data in CNN-based networks \cite{ActionRecognitionUsingDepthMaps&Postures}.


This study introduces a novel framework that integrates depth maps with RGB frames to enhance action recognition performance. Uniquely, we estimate depth maps from 2D frames instead of using depth cameras, making the approach more affordable and comparable to 2D video benchmarks. Our method uses the Side4Video (S4V) \cite{yao2023side4videospatialtemporalnetworkmemoryefficient} and VideoMamba \cite{li2024videomambastatespacemodel} frameworks, which employ CLIP \cite{radford2021learningtransferablevisualmodels} and VisionMamba \cite{zhu2024visionmambaefficientvisual} for extracting spatial features from RGB frames. We extended this by adding a separate branch for processing the estimated depth maps, followed by a late fusion of the extracted features. This approach captures actions' spatial and temporal dynamics more effectively by combining information from RGB and depth modalities.



\begin{figure}[tb]
  \centering
  \includegraphics[width=\textwidth,height=\textheight,keepaspectratio]{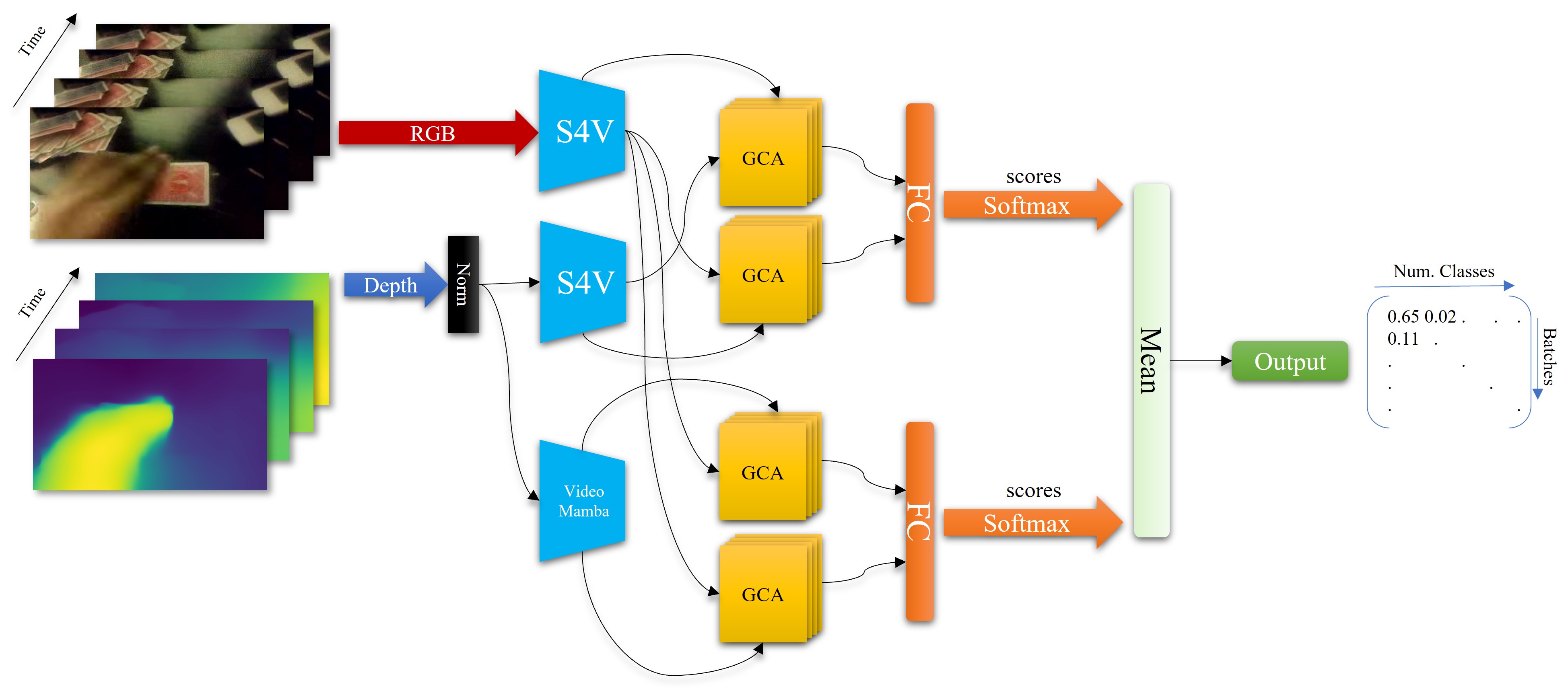}
  \caption{Overview of our method. We perform supervised action recognition using a fusion of RGB and depth map frames. We used S4V \cite{yao2023side4videospatialtemporalnetworkmemoryefficient} network for processing RGB and the same network plus VideoMamba \cite{li2024videomambastatespacemodel} to extract depth features. Gated cross-attention (GCA) has been used for fusing modalities, while mean operation has been selected for fusing models' scores.}
  \label{fig:FILS}
\end{figure}

\section{Enhanced Action Recognition with Depth and RGB}
\label{sec:method}
We introduce a framework for taking depth maps and RGB and analyzing their spatial and temporal features to recognize actions. Based on Figure \ref{fig:FILS}, our network's architecture consists of two major branches, one for processing RGB frames and another for depth maps. In this network, we use a late fusion manner to integrate extracted features using gated cross-attentions and a mean operation on streams' scores. In the next section, we describe the network architecture in more detail.

\noindent\textbf{RGB Encoder}
The network in this branch aims to extract both the temporal and spatial features of RGB frames. To achieve this, we employed the Side4Video (S4V) model \cite{yao2023side4videospatialtemporalnetworkmemoryefficient}, which uses the vision component of OpenCLIP \cite{radford2021learningtransferablevisualmodels} as the frame feature extractor. Side4Video uses a side network, where the trainable layers are added in parallel with the larger frozen model layers instead of in series. Specifically, in our model inspired by S4V \cite{yao2023side4videospatialtemporalnetworkmemoryefficient}, instead of adding new layers in series with the frozen CLIP \cite{radford2021learningtransferablevisualmodels} layers, we introduce side branches for the new layers. This approach ensures that backpropagation occurs only through the new layers, bypassing the frozen layers of CLIP. Consequently, this innovation allows for faster network training with more affordable resources.

\noindent\textbf{Depth Encoder} Although there are some depth maps datasets like \cite{shahroudy2016nturgbdlargescale}, we aim to use standard RGB video benchmarks. Therefore, we generate the depth directly from the 2D frames, using Midas \cite{ranftl2020robustmonoculardepthestimation} due to its balance between estimated depth maps' quality and generation speed. Since depth maps are in the standard RGB format to extract the depth map features, we employ an additional Side4Video (S4V) branch to encode the depth data. In addition, we propose a parallel encoder based on the mamba architecture. The incorporation of two sub-branches for depth processing in our network is motivated by the comparative strengths and unique perspectives of the VideoMamba \cite{li2024videomambastatespacemodel} and Side4Video (S4V) \cite{yao2023side4videospatialtemporalnetworkmemoryefficient} models. Their respective benchmark results (see Table~\ref{tab:SSV2}) show that VideoMamba is not as robust as S4V regarding overall performance. However, it offers a distinct understanding of the data. Since depth map features tend to be more sparse than RGB features, a less powerful but differently perceptive module, such as VideoMamba, is well-suited for this task. It is an architecture focused on sequence modelling to address sequence limitation and hardware consumption issues of transformers. We incorporate VideoMamba \cite{li2024videomambastatespacemodel} as an additional encoder processing the depth, resulting in our network comprising two depth encoder sub-branches. It uses VisionMamba \cite{zhu2024visionmambaefficientvisual},  to extract features of input images using Mamba, to process video frames to handle long-term high-quality videos that existing CNN and transformer-based networks might struggle with.

\noindent\textbf{Modality Fusion}
Following encoding the three modalities, we implement a late fusion policy to combine the data to contextualize features bidirectionally. This approach involves processing RGB and depth features through two sets of gated cross-attention (GCA) layers to contextualize each input based on another. It incorporates a gating mechanism that selectively filters and prioritizes the information passed between modalities. This gate acts like a finely tuned sieve, ensuring that only the most relevant data influences the model’s decision-making process. For the two layers, one uses the RGB features as the key and value, the depth as the query, and the key and value for the depth input and the RGB as the query. This process is repeated twice: once for fusing S4V depth maps and RGB features and once for fusing Mamba depth maps and S4V RGB features. A mean operation is applied on two model streams to combine their scores.

The overall model is trained with a cross-entropy loss function:
\begin{equation}
    \mathcal{L} = - \sum_{i=1}^{\mathit{Samples}} \sum_{c=1}^{\mathit{Classes}} y_{i,c} \log(\hat{y}_{i,c})
\end{equation}

where $y_{i,c}$ and $\hat{y}_{i,c}$ are true label and predicted probability for the i-th sample and c-th class

\section{Experiments}
\label{sec:exp} 
To assess the performance of our approach, we used  Something-Something V2 \cite{goyal2017something} (SSV2) to evaluate the action recognition downstream task.

\noindent{\textbf{Implementation Details.}}
To assess the performance of our approach, we used an input video clip with 8 frames each of size $224\times224$, and frames were chosen using static stride. The encoder that we are using is the ViT-B/16 architecture \cite{dosovitskiy2020image}, for which we used AdamW optimizer \cite{loshchilov2017decoupled}. Also, we set the batch size during inference to 26 as the optimum value.

\begin{table}[tb]
  \caption{ Out approach compared to other methods on SSV2. S4V* is our implementation}
    \centering
  \begin{tabular}{c |c c c c c}
    \toprule
    Method & Modality & Backbone & pre-train & Top-1\\
    \midrule
    SlowFast~\cite{feichtenhofer2019slowfast} & RGB  &ResNet101 & K400  & 63.1\\
    TSM~\cite{lin2019tsm} & RGB & ResNet50 & K400 & 63.4\\
    TimeSformer~\cite{bertasius2021space} &RGB &  ViT-L & IN-21K & 62.4\\
    Mformer~\cite{patrick2021keeping} & RGB & ViT-L & IN-21K+K400 & 68.1\\
    ViViT FE~\cite{arnab2021vivit}& RGB & ViT-L & IN-21k+K400 & 65.9\\ 
    VIMPAC~\cite{tan2021vimpac} & RGB & ViT-L & HowTo100M & 68.1\\     
    VideoMamba-S \cite{li2024videomambastatespacemodel} & RGB &- & IN-1K & 66.6 \\
    S4V*~\cite{yao2023side4videospatialtemporalnetworkmemoryefficient} & RGB & ViT-B & Clip-400M & 70.2 \\
    
    \midrule  
    Proposed & RGB+Depth(VideoMamba) & ViT-B & Clip-400M &  70.0\\
    Proposed & RGB+Depth & ViT-B & Clip-400M &  70.3\\
    \textbf{Proposed} & RGB+Depth+ Depth(VideoMamba) & ViT-B & Clip-400M &  \textbf{71.0}\\
  \bottomrule
  \end{tabular}

\label{tab:SSV2}
\end{table}

\noindent{\textbf{Results}}
Table \ref{tab:SSV2} presents the overall results and comparison with other methods. Our framework outperforms recent works in top-1 accuracy. Our full model \textbf{RGB+Depth+Depth(VideoMamba)} achieves 71.0\% accuracy; this is 4.4\% and 0.8\% greater than the individual data models VideoMamba-S and S4V that our approach is inspired by. The table also shows the fusion of depth and RGB, \textbf{RGB+Depth}, is 0.7\% lower than when we fuse the mamba logits too.

We also examined the impact of depth maps on various classes and found significant improvements in specific activities. For example, in class 145, \textit{Stuffing something into something}, the depth map helped it go up from 61.7\%
 to 72.9\%. Also, classes 65, 106, and 100, \textit{Pretending to be tearing something that is not tearable} and \textit{Putting something into something}, and \textit{Pushing something so that it slightly moves} achieved 5.6\%, and 7.5\%, and 7.1\% improvement, respectively, each by introducing depth maps. However, some classes, like 150, \textit{Tearing something just a little bit}, experienced a reduction of accuracy by 3.2\%, likely due to the lack of depth needed by this class for classification.

\section{Conclusion}
\label{sec:conclusion} 
In this work, we introduced a novel approach and architecture for enhancing action recognition by integrating depth maps. Our method estimates the depth maps of RGB frames, processes them through a separate branch from the RGB feature extractor, and subsequently fuses these features to understand the scene and actions comprehensively. We employed the Side4Video framework, leveraging the vision component of CLIP \cite{radford2021learningtransferablevisualmodels} to extract spatial features. Our approach demonstrated superior performance compared to our implementation of the Side4Video network. We benchmarked our method on the Something-Something V2 dataset, which provided a rigorous evaluation environment. For future work, we plan to apply this network to the EPIC-Kitchens dataset \cite{damen2022rescaling} to further validate its effectiveness across diverse and complex scenarios. This study underscores the potential of using depth maps alongside RGB features to improve action recognition, highlighting the importance of multi-modal data fusion in developing robust models for understanding intricate human-object interactions.

%
%

\section*{Acknowledgements}
Leverhulme Trust Research Project Grant RPG-2023-079 funded this work.

%
%
\bibliographystyle{splncs04}
\bibliography{tex/egbib}
\end{document}